\documentclass[conference]{IEEEtran}
\IEEEoverridecommandlockouts
\usepackage{cite}
\usepackage{amsmath,amssymb,amsfonts}
\usepackage{algorithmic}
\usepackage{graphicx}
\usepackage{textcomp}
\usepackage{xcolor}
\usepackage{multirow}

\usepackage{tabularx}
	
\usepackage[pages=some,placement=top, scale=1,
hshift=0,
vshift=-20,]{background}
\usepackage{lipsum}
\backgroundsetup{contents=This paper has been accepted for publication at 2022 IEEE International Conference on
Emerging Technologies and Factory Automation (ETFA),firstpage=true, angle=0,color=gray, opacity=1}

\def\BibTeX{{\rm B\kern-.05em{\sc i\kern-.025em b}\kern-.08em
    T\kern-.1667em\lower.7ex\hbox{E}\kern-.125emX}}

\makeatletter
\newcommand{\linebreakand}{%
  \end{@IEEEauthorhalign}
  \hfill\mbox{}\par
  \mbox{}\hfill\begin{@IEEEauthorhalign}
}
\makeatother

\usepackage{subcaption}
\begin{document}

\title{Improving safety in physical human-robot collaboration via deep metric learning\\
\thanks{This work was supported by ZHAW digital.}
}

\author{\IEEEauthorblockN{1\textsuperscript{st} Maryam Rezayati}
\IEEEauthorblockA{\textit{Institute of Mechatronics Systems} \\
\textit{Zurich University of Applied Sciences}\\
Winterthur, Switzerland \\
rzma@zhaw.ch}
\and
\IEEEauthorblockN{2\textsuperscript{nd} Grammatiki Zanni}
\IEEEauthorblockA{\textit{Department of Computer Science} \\
\textit{ETH Zurich}\\
Zurich, Switzerland \\
gzanni@student.ethz.ch}
\and
\IEEEauthorblockN{3\textsuperscript{rd} Ying Zaoshi}
\IEEEauthorblockA{\textit{Department of Computer Science} \\
\textit{ETH Zurich}\\
Zurich, Switzerland  \\
zaying@student.ethz.ch}
\and
\linebreakand %
\IEEEauthorblockN{4\textsuperscript{th} Davide Scaramuzza}
\IEEEauthorblockA{\textit{Robotics and Perception Group} \\
\textit{University of Zurich}\\
Zurich, Switzerland \\
sdavide@ifi.uzh.ch}
\and
\IEEEauthorblockN{5\textsuperscript{th} Hans Wernher van de Venn }
\IEEEauthorblockA{\textit{Institute of Mechatronics Systems} \\
\textit{Zurich University of Applied Sciences}\\
Winterthur, Switzerland \\
vhns@zhaw.ch}
}

\maketitle

\begin{abstract}
Direct physical interaction with robots is becoming increasingly important in flexible production scenarios, but robots without protective fences also pose a greater risk to the operator. In order to keep the risk potential low, relatively simple measures are prescribed for operation, such as stopping the robot if there is physical contact or if a safety distance 
is violated. 
Although human injuries can be largely avoided in this way, all such solutions have in common that real cooperation between humans and robots is hardly possible and therefore the advantages of working with such systems cannot develop its full potential. 
In human-robot collaboration scenarios, more sophisticated solutions are required that make it possible to adapt the robot's behavior to the operator and/or the current situation. Most importantly, during free robot movement, physical contact must be allowed for meaningful interaction and not recognized as a collision. However, here lies a key challenge for future systems: detecting human contact by using robot proprioception and machine learning algorithms.
This work uses the Deep Metric Learning (DML) approach to distinguish between non-contact robot movement, intentional contact aimed at physical human-robot interaction, and collision situations.
The achieved results are promising and show show that DML achieves 98.6\% accuracy, which is 4\% higher than the existing standards (i.e. a deep learning network trained without DML). 
It also indicates a promising generalization capability for easy portability to other robots (target robots) by detecting contact (distinguishing between contactless and intentional or accidental contact) without having to retrain the model with target robot data. 
\end{abstract}


\section{Introduction}

\begin{figure}[t]
    \centering
    \begin{subfigure}[t]{0.8\columnwidth}
        \centering
        \includegraphics[width=1\textwidth]{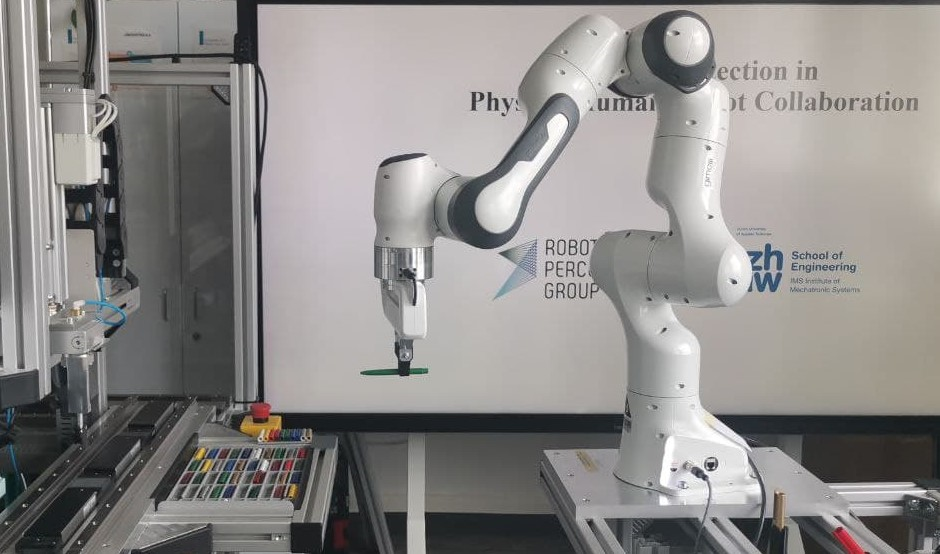}
        \caption{Contact-free motion}
    \end{subfigure}

    \begin{subfigure}[t]{0.5\columnwidth}
        \centering
        \includegraphics[width=.92\textwidth]{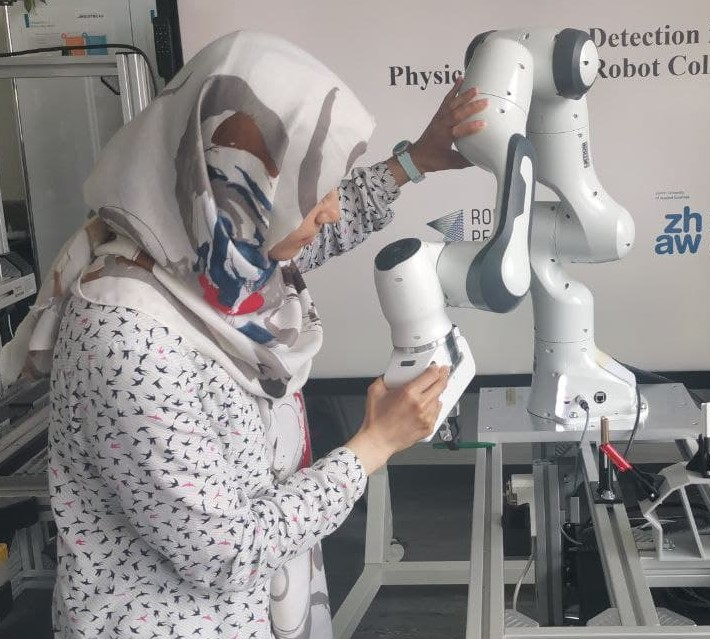}
        \caption{Intentional contact}
    \end{subfigure}%
    \begin{subfigure}[t]{0.5\columnwidth}
        \centering
        \includegraphics[width=1\textwidth]{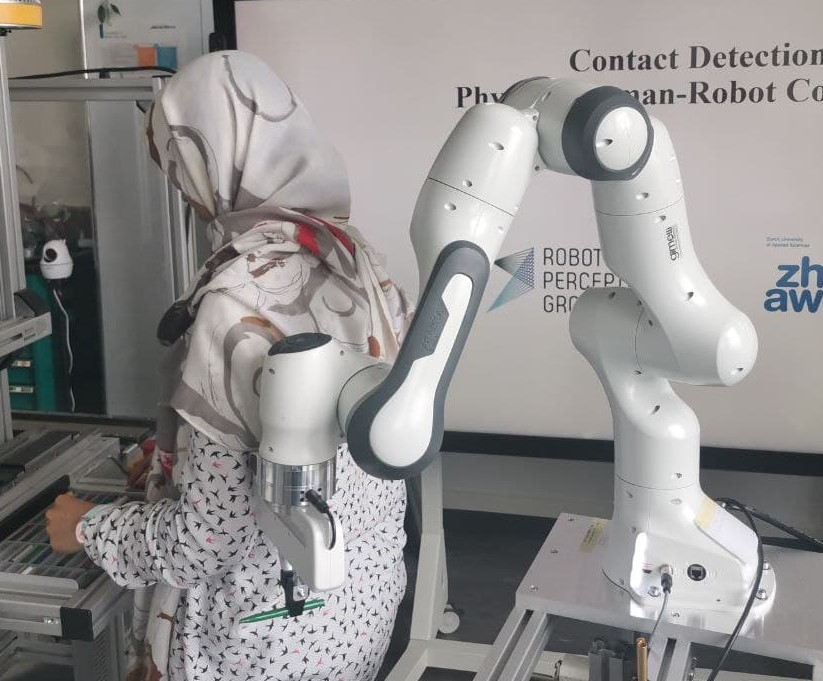}
        \caption{Incidental contact}
    \end{subfigure}
    \caption{Example of considered scenario in physical human-robot collaboration. Robots need to know whether it has a contact with human or not. Additionally, it is also imperative that robots can distinguish between intentional and accidental contacts.}
    \label{img:contactType}
\end{figure}

Over the past years, the international manufacturing industry has faced several serious challenges such as the shift from mass production towards customized products, accompanied by rapid changes in consumer trends, skilled labor shortage, and a lot more \cite{r1}. According to the World Robotics 2020 report, flexible industrial robot-based automation delivers the response to all of these challenges \cite{r1}. 

One solution to increase the rapid adaptation of automation sites is to combine human flexibility and cognitive abilities with robotic accuracy and endurance in terms of physical human-robot collaboration (pHRC). This combination decreases the human workload, improves the adaptability of the production line, and increases the acceptance of robots in a broader spectrum of human endeavors\cite{r2}. However, responsive collaboration and ensuring human safety are challenging with current robotic technology, as robots are endowed with no or very limited awareness of the immediate environment \cite{r2,r3,r4,r5} and cannot distinguish whether physical contact was explicitly initiated by the human operator, or whether it was caused by a collision situation (see Fig. \ref{img:contactType}).

To identify physical contact in industrial pHRC applications, there are two main approaches according to the current state of the art: model-based and data-based approaches. Model-based methods use mechanical and mathematical models of the robot dynamics to detect an unexpected contact. Some researchers propose sensor-less procedures which are based on a robot dynamics model\cite{r5, r6}, also through momentum observers \cite{r6,r7,r8,r9,r10 }, using extended state observers \cite{r11}, vibration analysis model \cite{r12}, finite-time disturbance observer \cite{r9}, and energy observer \cite{r10}.  Among these methods \cite{r5,r6,r7,r8,r9,r10,r11,r12}, momentum observers have better performance in real situations than other types. However, they require the exact dynamic parameters of the robot \cite{r14}, which are difficult to determine in real-world applications. Accordingly, data-driven approaches offer a range of satisfactory alternatives as they do not rely on the mechanical model but on the data collected from the robot. Therefore, data-driven approaches such as being used in machine learning, deep learning and AI, have recently been prevalent in classifying contact types on the basis of the robot sensors’ data.

Several recent studies have been conducted about implementing AI in the area of contact detection. In one method, model-driven approaches’ performance is enhanced by incorporating AI algorithms\cite{r15,r16}. A deep neural network designed by Anvaripour and Saif categorizes the output of a sliding mode observer external force estimator, into contact and non-contact classes which eliminates the difficulty of threshold determinations \cite{r15}. Park et al. evaluate the performance of a convolutional neural network (CNN) and a support vector machine (SVM) in classifying the output of another external torque estimator, a so called momentum observer \cite{r16}. According to their results, the CNN-based algorithm performs better with big training data, while the SVM-based algorithm responds better with small training data \cite{r16}. As an alternative, some researchers use robot sensor data as an input to AI algorithms without including robots’ dynamic models \cite{r17,r18,r19,r20}. Popov et al. attempted to detect robot collisions and classify them into soft and hard collision using multi perceptron neural networks (MLP) on the basis of internal joint torque and encoder measurements \cite{r17}. They demonstrated that neural network solutions can decrease false negative and false positive errors significantly in comparison to threshold solutions \cite{r17}. Czubenko and Kowalczuk designed external torque estimators using various neural networks such as auto-regressive, recurrent neural network (RNN), CNN long short-term memory (CNN-LSTM), and mixed convolutional LSTM network\cite{r19}. The outcome of their work indicates that a mixed convolutional LSTM network was the most effective solution. 

Since few applications of AI have been studied in the contact detection field to increase robot perception, it is still necessary to explore other algorithms that are both relevant and technically feasible in order to achieve a sustainable safety system. In addition, the collection of human-robot collision datasets is typically less meaningful due to the fact that the collision was deliberately induced to ensure the safety of humans during data collection\cite{r3}. We hypothesis that one paradigm to address this issue could be increasing the explainability of the network pipeline. For this goal, we aim to do the classification task in two steps;
\begin{itemize}
    \item First step: mapping the input features to an explainable space where similar samples have small distance form each other and far enough from dissimilar samples
    \item Second step: classifying the mapped input features.
\end{itemize}
A powerful machine learning tool to develop such a mapping space is Deep Metric Learning (DML)
DML mimics the human cognitive process to detect similarities between objects. In this case, the objects are the contact patterns between human and robot. DML now uses deep learning to estimate pairwise similarity / distance between the occurring events and a reference pattern. The better the similarity, the more closely the measured signal corresponds to the respective reference signal, or vice versa, the further apart the signal patterns are from one another, the less good is their match. The definition of good similarity measure (metric) is task dependent.
Hence, we favor the DML technique as it can boost the model performance and we demonstrate the effectiveness of this idea using different loss functions: 
\begin{itemize}
    \item Pairwise loss function with Siamese Network, a previous DML method, which judges whether the two inputs belong to the same class or not.
    \item Magnet loss function, which can adaptively sculpt its representation space by autonomously identifying and respecting intra-class variation and inter-class similarity \cite{r21}.
    \item Triplet loss function which enhances the capability of DML by using Triplet network and dividing the samples into anchor, positive, and negative \cite{r22}. There are three kinds of Triplet loss function as follows: the easy triplet loss function, semi-hard triplet loss function and hard triplet loss function. Hereby, the semi-hard triplet loss function is chosen for the good convergence properties and additionally, for the relatively proper distances between the Anchor and the Positive and Negative pair.
\end{itemize}
 The aim of this study is first to understand how DML is used in contact recognition/perception systems and to compare it to a baseline built up with the same network architectures and trained with a common loss function, sparse categorical cross-entropy loss. Secondly, we want to compare the performance of different DML loss functions in improving the contact perception system. As a third goal, we investigate our model generality for the application with other robots.

\textbf{Our contributions} are briefly as follows:
\begin{itemize}
    \item We propose a DML-based method for identifying human contacts with a robot using the robot's proprioception data and through investigating a range of network architectures and DML loss functions. The presented method improves the performance of contact perception which enhances human safety in a shared workspace. 
    \item The presented approach was implemented on a real robot and indicates that it not only outperformes the state of the art in offline testing scenarios but also improves contact recognition in real-time deployment.
    \item DML-based trained architectures demonstrate 
    outstanding generalization  in detecting physical contact (no matter which type of contact) to other robots with the same type but different calibration, sensor noise, uncertainty, and robot trajectory \textbf{without having any data from the target robot}. Such a generalization has not been observed and reported in any similar study to the best of our knowledge.
\end{itemize}

\begin{figure*}[t]
    \begin{center}
        \includegraphics[width=1\textwidth]{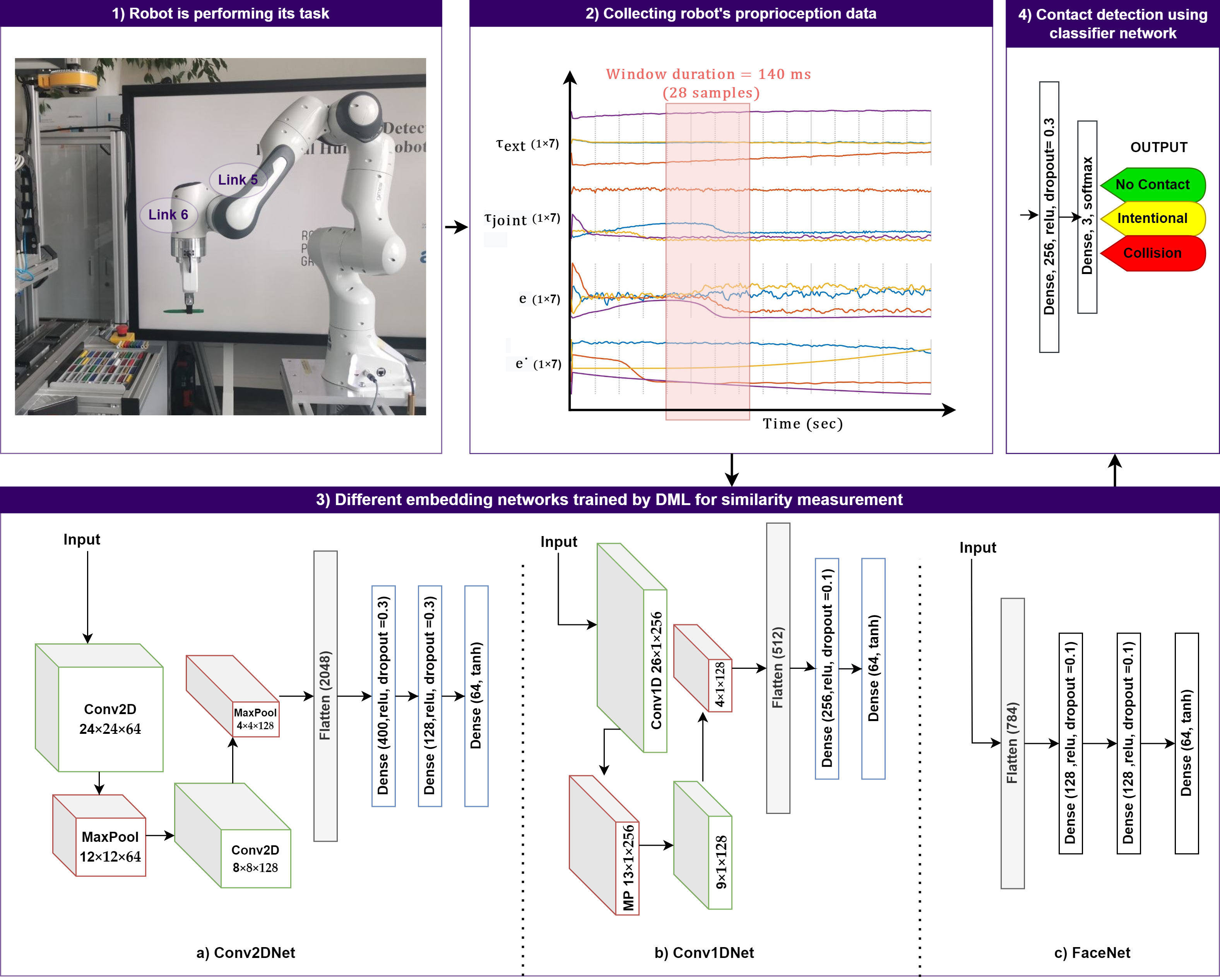}
        \caption{Proposed approach. Data is collected from a Franka Panda robotics arm with sampling rate of 200Hz. A time window containing 28 samples with 28  features are used as the input for our ML algorithm. Our model consists of two parts; Embedding network and contact detection classifier. The performance of three different embedding networks trained by three different DML loss functions is compared in the design of the contact perception system.}
        \label{img:proposed_approch}
    \end{center}
\end{figure*}

\section{Material and Methods}

\subsection{Our Approach}
The proposed pipeline of our contact perception system is shown in Fig. \ref{img:proposed_approch} which is designed inline with our goals of this study. It contains two main modules, an embedding network, and a classifier network. The embedding network is trained using DML and it maps the input data to an embedding space. The mapped input data then will be classified using a classifier network to specify whether there is a contact or what the nature of that contact is. 

For conducting our experiments, a set of network architectures (please refer to section D) 
are used to be trained by our metrics. Each training is done several times with a common set of different random initialisation. Moreover, in order to have a better comparison about embedding networks, a simple deep network, shown in Fig. \ref{img:proposed_approch}, will be used for contact detection which classifies the output of the embedding networks into three classes: non-contact, intentional contact, and collision. After comparing trained models with different loss functions and network architectures on the test set, the best model is implemented on the source robot (the robot used for data collection) to test the real-time deployment result. Lastly, we test this model on another robot, the targeting robot, which is of the same type as source robot, without retraining the model to test the generalizability of our approach. This is a test to show that the generalization is possible in principle. Further research will be undertaken to also show generalization for different types of robot.

\subsection{Robot Platform}
Two Franka Emika Panda robotic arms are used throughout the entire project, recognized as agile and sensitive collaborative robots. We name one of them "the source robot", which is used for collecting the dataset and testing models in real-time implementation, and another one as "the target robot" to test the generalization performance of our method to another robot (of the same type) in a real-time implementation without retraining the model. In other words, the model is trained with the data of source robot and will be deployed and tested on the target robot. 

The following paragraph gives a summary of the main characteristics of this kind of robot; the weight of the robot is approximately 18 kg, and it can carry a maximum payload of three kilograms. An arm and a hand are the two main components of the robot. The arm is a 7DOF manipulator with revolute joints solely. Each joint is equipped with an encoder and an external torque sensor (13 bits resolution). Due to a force controller, the gripper is able to grasp objects securely. Furthermore, it is possible to control the robot with either external or internal torque controllers at a frequency of 1 kHz.

\subsection{Dataset collection and input layer}

We adopted a collision detection dataset from our previous study \cite{r3}, which is publicly available in \cite{r23}. The dataset is collected from a real-world pick-and-place application in the smart pen manufacturing demonstrator of our laboratory \cite{r29}. The task of the robot is to move pen components from a tray to the demonstrator pen parts container with a joint velocity less than 0.7 rad/s (see Fig.\ref{img:contactType}) while the operator can help the robot find the correct position of the pen part in the container by physically guiding it at individual joints, shown in Fig.\ref{img:contactType}.b (intentional contact). The robot data in the five different states is collected and labeled manually as follows:
\begin{itemize}
    \item Non-contact: a robot preforms its task without having any contact to human operators or other objects.
    \item Intentional Link5: while the robot is doing its task, the operator grasps the fifth link of the robot to interact with it.
    \item Intentional Link6: while the robot is doing its task, the operator grasps the sixth link of the robot to interact with it.
    \item Incidental Link5: the fifth link of the robot collides with a human body.
    \item Incidental Link6: the sixth link of the robot collides with a human body.
\end{itemize}
The dataset includes 2146 samples while each sample is a time-series window containing 1-second time-lapse robot sensor data; Motor torque, external applied torque, joint position, and velocity. The data is collected with a sampling rate of 200 Hz. As pen parts had negligible weights (less than 20 gr), we did not collect data for different payloads. Additionally, the collision data just gathered from link 5 and link 6 since it is the first try to distinguish between intentional and incidental contacts, but later we want to expand it to all joints for both incidental and intentional contact. Furthermore, since the focus of our study is on human safety, collisions between the robot  and other objects are not considered in our trials.

To preprocess the datasets for this study, sensor data is categorized with associated labels: non-contact, intentional and collision. The input vector represents a time series of robot data and is represented as (see Fig.\ref{img:proposed_approch}):

\begin{equation}
    x=\begin{bmatrix}
    \mathbf{\tau_J^0} & \mathbf{\tau_{ext}^0} & \mathbf{e^0} & \mathbf{\dot{e}^0}\\
    \mathbf{\tau_J^1} & \mathbf{\tau_{ext}^1} & \mathbf{e^1} & \mathbf{\dot{e}^1}\\
    \vdots &\dotsi &\dotsi & \dotsi \\
    \mathbf{\tau_J^{28}} & \mathbf{\tau_{ext}^{28}} & \mathbf{e^{28}} & \mathbf{\dot{e}^{28}}\\
    \end{bmatrix}
\end{equation}
and 
\begin{equation}
    \mathbf{\tau_J^i}=\begin{bmatrix}
    \tau_{J_1}^i &\tau_{J_2}^i&\tau_{J_3}^i&\tau_{J_4}^i &\tau_{J_5}^i&\tau_{J_6}^i&\tau_{J_7}^i 
    \end{bmatrix}
\end{equation}
\begin{equation}
    \mathbf{\tau_{ext}^i}=\begin{bmatrix}
    \tau_{ext_1}^i&\tau_{ext_2}^i&\dotsi&\tau_{ext_6}^i &\tau_{ext_7}^i
    \end{bmatrix}
\end{equation}
\begin{equation}
    \mathbf{e^i}=\begin{bmatrix}
    e_1^i & e_2^i & e_3^i & e_4^i & e_5^i & e_6^i & e_7^i
    \end{bmatrix}
\end{equation}
\begin{equation}
    \mathbf{\dot{e}^i}=\begin{bmatrix}
    \dot{e}_1^i & \dot{e}_2^i & \dot{e}_3^i & \dot{e}_4^i & \dot{e}_5^i & \dot{e}_6^i & \dot{e}_7^i
    \end{bmatrix}
\end{equation}
where $\tau_J$, $\tau_{ext}$, e, and $\dot{e}$ indicate joint torque, external torque, joint position error, and joint velocity error, respectively. The input vector is 28×28 and indicates 140ms time frame data which is less than reaction time to haptic stimuli for humans (155ms \cite{r28}). In our trials, shorter time frame lead to inappropriate detection accuracy while longer window increase computational costs. The time frame range was from 1500ms to 100 ms. It might also depend on robot speed which would need to be investigated, however this aspect was out of scope of this paper.
\subsection{Embedding Networks Architecture}
Specifying embedding networks, we examined different architectures and selected three different deep networks, shown in Fig. \ref{img:proposed_approch}, which were better than our other trials for being trained by our metrics; Conv2DNet, Conv1DNet, and FaceNet. In our experiments, embedding space with size of $64\times1$ was suitable for our task. 

\textbf{Conv2DNet} is a 2D convolutional neural network with 8 layers as follows
\begin{itemize}
    \item 2D convolutional layer with 64 filters, kernel size 4
    \item 2D Max pooling layer with pool size 2
    \item 2D convolutional layer with 128 filters, kernel size 4
    \item 2D Max pooling layer with pool size 2
    \item Flatten layer
    \item Dense layer with 400 neurons
    \item Dense layer with 128 neurons
    \item Dense layer with 64 neurons
\end{itemize}

\textbf{Conv1DNet} is a 1D convolutional neural network with 7 layers as follows
\begin{itemize}
    \item 1D convolutional layer with 256 neurons, kernel size 3
    \item 1D Max pooling layer with pool size 2
    \item 1D convolutional layer with 128 neurons, kernel size 4
    \item 1D Max pooling layer with pool size 2
    \item Flatten layer
    \item Dense layer with 256 neurons
    \item Dense layer with 64 neurons
\end{itemize}
\textbf{FaceNet} is a multi-layer perceptron network (MLP) \cite{r24}. It has a flatten layer and three dense layers with 128, 128, 64 neurons respectively. The dropout for the hidden layers are set to 0.1. 
All network architectures are constructed with an input dimension of 28x28 and an output dimension of 64x1.

\subsection{Pairwise loss function}

\begin{figure}[t]
    \centering
    \begin{subfigure}[t]{\columnwidth}
        \centering
        \includegraphics[width=1\textwidth]{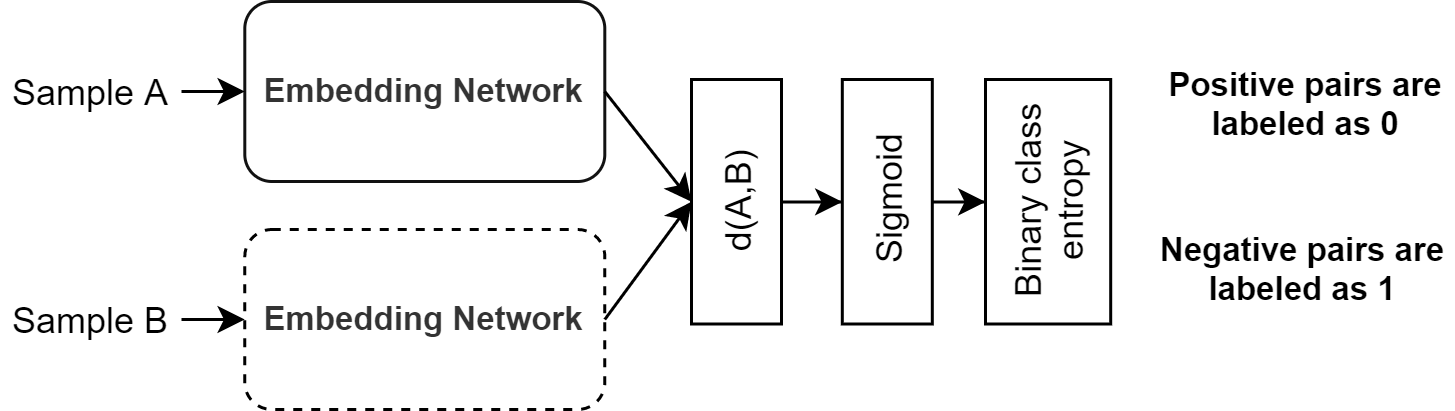}
        \caption{Pairwise loss function}
    \end{subfigure}\\
    \begin{subfigure}[t]{\columnwidth}
        \centering
        \includegraphics[width=0.8\textwidth]{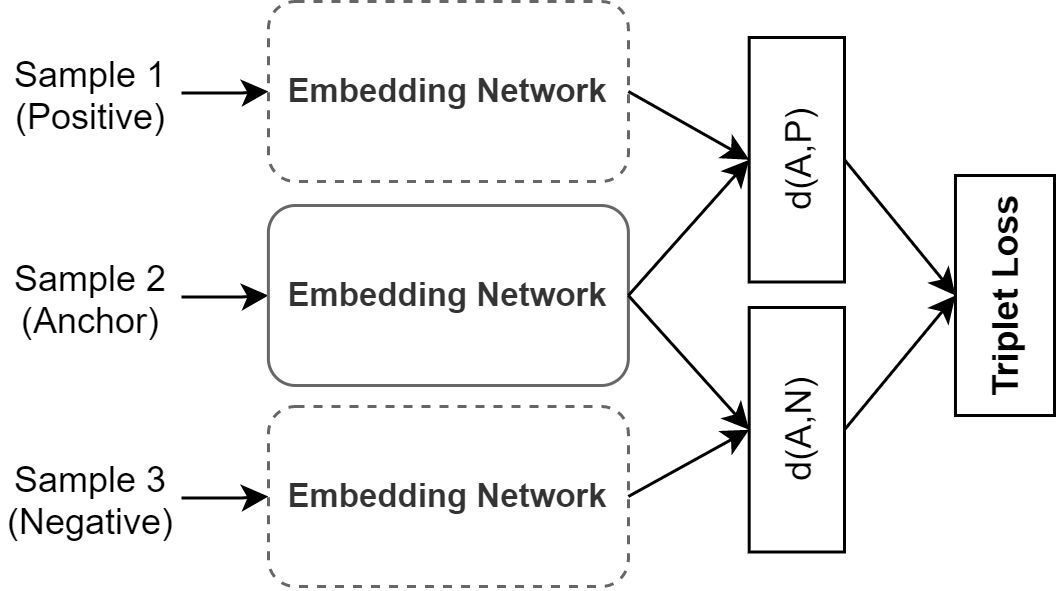}
        \caption{Triplet loss function}
    \end{subfigure}\\
    \begin{subfigure}[t]{\columnwidth}
        \centering
        \includegraphics[width=0.9\textwidth]{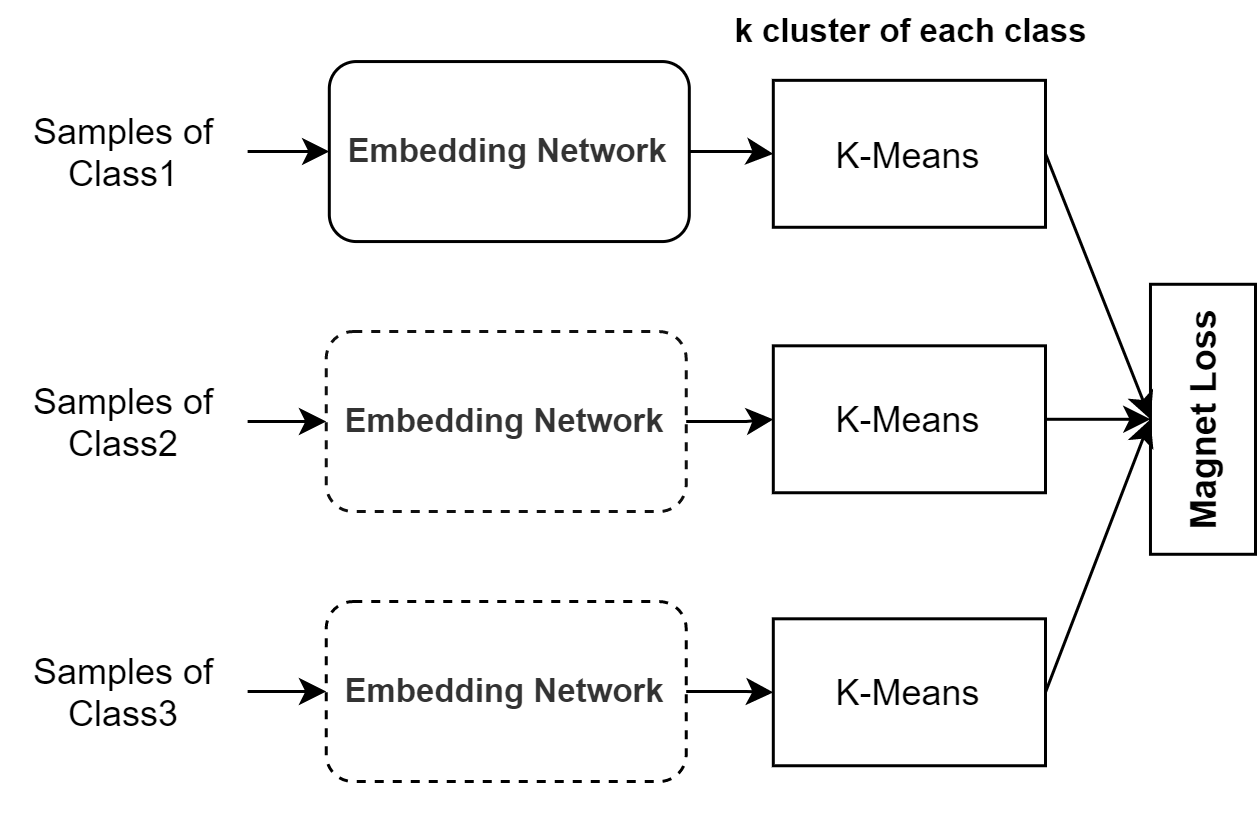}
        \caption{Magnet loss function}
    \end{subfigure}
    \caption{Training approach of embedding networks with a) pairwise b) triplet c) magnet loss functions}
    \label{img:trainingLoss}
\end{figure}

Pairwise learning refers to the fact that we learn a metric embedding by means of pairs of samples. The idea is, to generate a dataset of positive, or similar, pairs and negative, or dissimilar, pairs. In the context of this paper, positive pairs are generated by drawing random pairs of samples from the same class, whilst negative pairs are generated by drawing random samples from different classes. We use a Siamese network, a pair of identical networks, whose weights are tied, that is, weights will evolve together as the training progresses. We cast the pairwise metric learning task into a classification task based upon the distance of the pairs of points in the latent embedding. More precisely:
\begin{itemize}
    \item A pair of points is processed in parallel by the two identical networks, this allows us to obtain a latent representation for both samples, or, two feature vectors in the embedding we aim at learning.
    \item Then the we take the L1 distance between these two feature vectors and this is compressed back into the range [0, 1] by applying the sigmoid function.
    \item Finally, the binary cross entropy loss is applied.
\end{itemize}
See Fig. \ref{img:trainingLoss}(a) for a visual representation.
\subsection{Triplet Loss Function}
This network encodes the pair distances between each positive and negative sample against the anchor sample. The Anchor and the Positive are samples describing the same entity, while the Anchor and the Negative are not. Therefore, it can get the distance between samples and decide which ones are more similar \cite{r24,r25}. The overall network structure of the triplet network with the triplet loss function is shown in Fig. \ref{img:trainingLoss}(b). We define the Model such that the Triplet Loss function receives all the embedding from each batch, as well as their corresponding labels (used for determining the best triplet-pairs). This is done by defining an input layer for the labels and then concatenating it to the embedding. The loss function is defined as:
\begin{equation}
    L = max(d(A,P)-d(A,N)+margin,0)    
\end{equation}
where $d(A,P)$ and $d(A,N)$ are the Euclidean distances between the Anchor and the Positive and Negative pairs, margin is a parameter making the network learn a specific distance between positive and negative samples by using the anchor. By setting a proper value for margin, we can make the convergence easier, meanwhile, ensuring that $d(A,N)$ is enough larger than $d(A,P)$.
\subsection{Magnet Loss Function}
Magnet learning, shown in Fig. \ref{img:trainingLoss}(c), is structured similarly to the Triplet while in the Magnet loss function, an explicit model of different classes’ distribution is maintained instead of penalizing just pair or triplet samples. The magnet loss function is defined as \cite{r22}: 

\begin{equation}
    L=\frac{1}{N}\sum_{n=1}^{N} \bigg\{ -\log(
    \frac
    {e^{-\frac{1}{2\sigma^2}\|r_n+\mu(r_n)\|_2^2-\alpha}}
    {\sum_{c \neq C(r_n)} \sum_{k=1}^K e^{-\frac{1}{2\sigma^2}\|r_n+\mu_k^c\|_2^2}}
    )\bigg\}_+
\end{equation}

\begin{equation}
    \sigma^2 = \frac{1}{N-1} \sum_{r \in D} \|r-\mu(r)\|
\end{equation}
where {.}+ is the hinge function, $\alpha \in$ R is a scalar, $r_n$ denotes $x_n$ in the embedding space, $C(r_n)$ is class representation of $r_n$, $k$ is the cluster number where each class has $K$ clusters obtained by K-Means algorithm. $\sigma^2$ is the variance of all examples away from their respective class centers. For Magnet Loss, an entire local neighborhood of nearest clusters is retrieved iteratively, and their overlaps are penalized. 
\subsection{Real\textendash Time Implementation}
\begin{figure}
    \centering
    \includegraphics[width=0.48\textwidth]{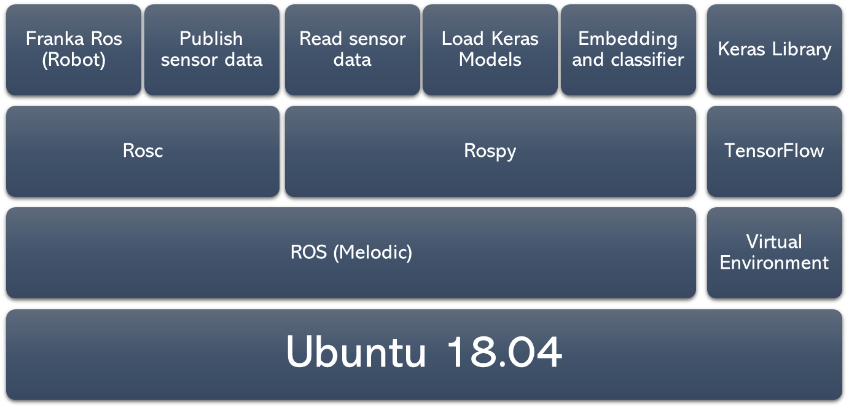}
    \caption{Software architecture of real-time implementation}
    \label{img:implementation}
\end{figure}

The real-time interface for reading the data and implementing the trained network on the robots was provided by the Robotics Operating System (ROS) on Ubuntu 18.04 LTS with real-time kernel.  This is our working environment for the time being but also other versions can be used, there is no specific need for Ubuntu 18. As shown in Fig. \ref{img:implementation}, ROS uses a C-library, rosc, to communicate with the robot and uses a python-library, rospy, to run the embedding and classifier networks in real-time. The computer connected to the robot controller has 16GB RAM, Intel Xeon CPU, and NIVIDIA Quadro K4200 GPU. 

\section{Experimental Results}

\begin{figure}[t]
    \centering
    \begin{subfigure}[t]{0.47\columnwidth}
        \centering
        \frame{\includegraphics[width=1\textwidth]{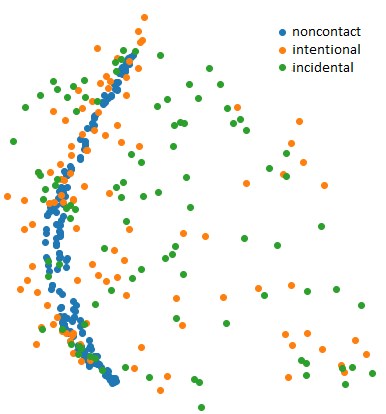}}
        \caption{}
    \end{subfigure}%
    \begin{subfigure}[t]{0.48\columnwidth}
        \centering
        \frame{\includegraphics[width=1\textwidth]{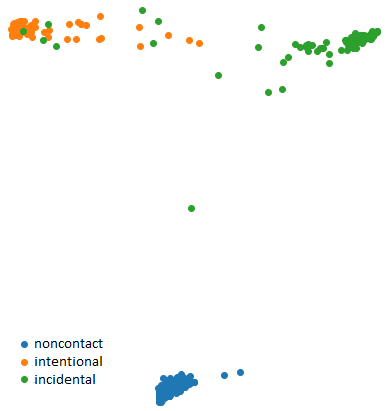}}
        \caption{}
    \end{subfigure}
    
    \begin{subfigure}[t]{0.48\columnwidth}
        \centering
        \frame{\includegraphics[width=1\textwidth]{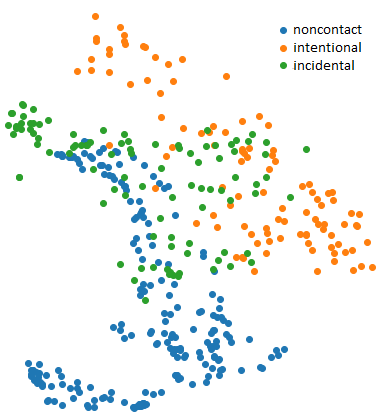}}
        \caption{}
    \end{subfigure}%
    \begin{subfigure}[t]{0.48\columnwidth}
        \centering
        \frame{\includegraphics[width=1\textwidth]{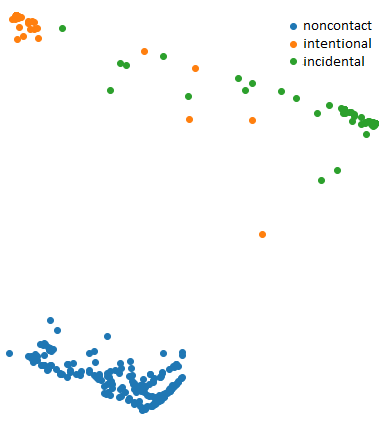}}
        \caption{}
    \end{subfigure}
    \caption{Embedding visualization of Conv1DNet a) before training b) trained with Pairwise loss function c) trained with Magnet loss function d) trained with Triplet loss function}
    \label{img:embeddingSpace}
\end{figure}

\begin{table}
    \centering
    \caption{Performance of embedding networks trained by Siamese, Magnet, and Triplet loss functions}
    \begin{tabular}{m{1cm} m{1.95cm} m{1.95cm} m{1.95cm}}
        \hline
         &FaceNet &Conv2DNet &Conv1DNet \\
        \hline
    \end{tabular}
    \begin{tabular}{m{1cm}m{0.75cm} m{0.75cm} m{0.75cm} m{0.75cm} m{0.75cm} m{0.75cm}}
        \hline
         &Loss &Acc &Loss &Acc &Loss &Acc \\
        \hline
    \end{tabular}
    \begin{tabular}{m{1cm}|m{0.75cm} m{0.75cm} m{0.75cm} m{0.75cm} m{0.75cm} m{0.75cm}}
        Pairwise & 0.3267 &	0.8298 & 0.4311 & 0.7879 & 0.3914 & \textbf{0.9907}\\
        Magnet & 0.2615 & 0.9464 & 0.1506 & 0.9534&	0.2555 & 0.9674\\
        Triplet & 0.1456 & \textbf{0.9767} & 0.1597 & \textbf{0.9767} & 0.1121 & 0.9860\\
        Baseline$^\ast$ & 0.1324 &	0.9487 & 0.1604 & 0.9487 & 0.1375 & 0.9417\\
        \\
    \end{tabular} \\
    $\ast$ Baseline built up with the same network architectures and trained with a common loss function, sparse categorical cross-entropy loss.
    \label{tab:trainResult}
\end{table}

\subsection{Contact Perception System using DML}
Detailed explanations of the required tests before evaluating the final model on the Panda robot are provided in this section. In the first step, we will examine the performance of DML in the classification of the contact detection dataset. For doing so, three different networks are trained with the aforementioned loss functions and compared with the baseline. As baseline, we choose to train networks with the same architecture but without using DML loss function. 80\% of the dataset is used for training and 20\% for testing the trained models. The results in Table \ref{tab:trainResult} represent the accuracy and loss of classifications based on the testing set. For all networks, the highest accuracy belongs to DML (pairwise, magnet or triplet) which shows that DML outperforms the baseline accuracy. This result indicates that DML can help improving the accuracy of a contact perception system.
Another conclusion is that the Conv1DNet delivers a better result in comparison to the other networks for all loss functions in terms of loss and accuracy. A high level of accuracy is also achieved for Conv2DNet, unless it is trained with the pairwise Loss function.
According to Table \ref{tab:trainResult}, the accuracy of models trained by pairwise loss function varies significantly by changing network architecture whereas it varies less than 2\% for other loss functions. 
As shown in Fig. \ref{img:embeddingSpace}, the embedding visualization\footnote{Visualization is done using principal component analysis (PCA) to reduce the dimension of the embedding space\cite{r29}} of the best trained network, Conv1DNet, demonstrates that the pairwise and triplet loss functions led to similar embedding space and accuracy.
The confusion matrix in Table
\ref{tab:trainpairwise}, indicates both networks distinguish ideally the non-contact class, while the intentional and collision classes are not distinctly separated. Nevertheless, the triplet learning has less false positive and false negative errors than the pairwise one, even though the contact classes are still not perfectly separated.

\subsection{Generalization and Real-Time Implementation}

Table \ref{tab:robots} shows the performance of the best model, Conv1DNet trained by triplet loss function, in the real-time implementation on the source and target robots. This result is described in the terms of detection failures and false alarms which are false negatives and false positives of a
classification task respectively. Considering the non-contact class, detection failures and false alarms are zero on both robots. This outcome indicates that the system not only can detect the physical contact on the source robot perfectly but also has a great generalization ability to correctly detect the contact on the target robot. Regarding the contact type recognition (considering incidental and intentional contact classes), the result shows that the model has less detection failures in the collision class than in the intentional contact class which is of interest for human safety. However, the model generalization for contact type recognition is still challenging as the target robot has different calibration, sensor noise, uncertainty, and robot trajectory than the source robot.  This leads to a shift in the robot data distribution  which results in the observed weakness of our model. We will place more emphasis on working with domain adaptation techniques in our future research.

\begin{table}
    \centering
    \caption{Confusion matrix of Conv1DNet trained by Pairwise and Triplet loss function}
    \begin{tabular}{c}
        \\
        \\
        \text{Predicted Label}\\
    \end{tabular}
    \begin{tabular}{c|c c c|c c c}
        \multicolumn{7}{c}{True Label}\\
        \multicolumn{4}{c}{Pairwise} &\multicolumn{3}{c}{Triplet}\\
          &C1 &C2 &C3 &C1 &C2 &C3\\
        \hline
         C1 &228 &0 &0 &228 &0 &0 \\
         C2 &0 &100 &10 &0 &102 &4\\
         C3 &0 &4 &87 &0 &2 &93\\
         \multicolumn{7}{c}{}\\
    \end{tabular}
    \text{C1: noncontact, C2: Intentional, C3: Collision}
    \label{tab:trainpairwise}
\end{table}

\begin{table}
    \centering
    \caption{Performance of Conv1DNet trained by Triplet loss function in the real-time implementation on source and target robots}
    \begin{tabular}{c|c c c|c c c}
        \multicolumn{1}{c}{}&\multicolumn{3}{c}{Source Robot} &\multicolumn{3}{c}{Target Robot}\\
          &C1 &C2 &C3 &C1 &C2 &C3\\
        \hline
         Detection failures &0/447 &6/19 &1/11 &0/612 &17/34 &13/30 \\
         False alarms &0/447 &1/19 &6/11 &0/612 &13/34 &17/30\\
         \multicolumn{7}{c}{}\\
    \end{tabular}
    \text{C1: noncontact, C2: Intentional, C3: Collision}
    \label{tab:robots}
\end{table}

\section{Discussion}

The distinction between incidental contact and intentional contact is of crucial importance for human-robot collaboration, since it can elevate the physical interaction with the robot for humans to a level that corresponds to the natural interaction between humans. This favors a more intuitive, responsive and proactive form of collaboration, which is usually not possible today due to the necessary safety precautions such as low robot speed and emergency stop when a contact is detected. Ultimately, the key factor for safe and efficient HRC in a variety of applications from industrial manufacturing and logistics to healthcare assistance systems is understanding human intent. The great benefit lies in the goal of significantly increasing the safety of cooperation and at the same time increasing efficiency. Proactive behavior enables simpler and more efficient safety concepts and will definitively advance applications in all areas.

When comparing the baseline networks with the three DML networks, it is obvious that DML in some cases significantly improves the training results. The reason for this improvement could be due to the design of our pipeline. We favor the DML to train the embedding network to map the input data into an explainable space in which the samples of one class are the minimum distance from each other and the maximum distance from other classes. This makes the classification task easier since the input features are less complicated. We plan to further investigate this effect more carefully by changing the modules of this pipeline in our future research to find the best approach for further optimization.

The Conv1DNet trained by triplet loss function is generally the best implementation of DML on this dataset. In addition, the Pairwise loss function provides the best in accuracy in Conv1DNet, which indicates that it may deal better with lower complexity networks. Obviously, further experiments must be carried out in order to find a valid explanation for this behavior.

Amongst the state-of-the-art contact detection approaches, there are three similar sources investigating collision detection using 1D-CNN. The authors of \cite{r3} compared these approaches; CollisionNet \cite{r26}, FMA \cite{r15}, and CDN \cite{r3} where the accuracy was 88\%, 90\%, and 96\% respectively, featuring a detection delay of 200ms \cite{r15} and 80ms \cite{r3}. In our case, the best model reached 98\% with 50ms detection delay. Another noteworthy advantage is, that in \cite{r3} the time frame window is 1 second, while in \cite{r28} the reaction time to haptic stimuli for humans is reported to about 155ms. With our design, we were able to achieve a time frame window of 140 ms, which is more agile for safety monitoring systems in human-robot collaboration applications.

\section{Conclusion and Outlook}
In order to detect individuals and protect them during physical human-robot collaboration, robots need to have a higher level of perception. In this paper, a deep metric learning approach is employed to improve robot comprehension of external contacts. To achieve this aim, three different deep networks - FaceNet, Conv2DNet, and Conv1DNet - are trained by three different DML loss functions—Pairwise, Magnet, and Triplet— and then compared to the baselines. In conclusion, Conv1DNet trained by triplet loss function has the most accurate model in terms of accuracy (0.9860) and loss (0.1121). Furthermore, it demonstrated promising performance in detecting real-time contact, while it was not able to clearly distinguish the intentional and collision contacts. Lastly, the model deployment on the target robot indicates that it has a prominent generalization power in detecting physical contact regardless its type. However, further research is required to improve the model accuracy in contact type detection.

We have a limitation in that our study only investigated robot speeds under $0.7 rad/s$ joint velocity, and we should investigate this approach for higher speeds since it has a direct effect on increasing the productivity of HRC systems.
In the future, we intend to increase the generality of our models using domain adaptation. On the other hand, the DML approach has not been proven yet as the best method in this robotic application, given that other deep cluster approaches are still not tested and compared, for instance, DBSCAN. For future work, more experiments and research will be conducted on the two above-mentioned topics.

\vspace{12pt}

\end{document}